\documentclass[twoside,11pt]{article}

\usepackage[accepted]{melba}

\usepackage{amssymb}
\usepackage{xcolor}

\usepackage{gensymb}

\usepackage{graphicx}
\usepackage{enumitem,kantlipsum}

\usepackage{color}

\melbaid{2023:014}  
\doi{https://doi.org/10.59275/j.melba.2023-3d9d}
\melbaauthors{Singh, Jasra, Mouhammed, Dadheech, Ray and Shapiro} 
\volume{2}
\firstpageno{390}  
\melbayear{2023}  
\datesubmitted{05/2023}  
\datepublished{10/2023}  

\makeatletter
\def\cat@comma@active{\catcode`\,12}%
\makeatother

\ShortHeadings{Towards Early Prediction of Human iPSC Reprogramming Success}{Singh, Jasra, Mouhammed, Dadheech, Ray and Shapiro}

\title{Towards Early Prediction of Human iPSC Reprogramming Success}

\author{\name Abhineet Singh\orcid{0000-0002-5624-8377} \email asingh1@ualberta.ca
	\\ \addr Department of Computing Science, University of Alberta
	\AND
	\name Ila Jasra\orcid{0000-0002-2842-8565} \email jasra@ualberta.ca
	\\ \addr Alberta Diabetes Institute, University of Alberta
	\AND
	\name Omar Mouhammed\orcid{0000-0003-3537-9420} \email mouhamme@ualberta.ca
	\\ \addr Alberta Diabetes Institute, University of Alberta
	\AND
	\name Nidheesh Dadheech\orcid{0000-0002-7155-0486} \email dadheech@ualberta.ca
	\\ \addr Alberta Diabetes Institute, University of Alberta
	\AND
	\name Nilanjan Ray\orcid{0000-0002-7588-5400} \email nray1@ualberta.ca
	\\ \addr Department of Computing Science, University of Alberta
	\AND
	\name James Shapiro\orcid{0000-0002-6215-0990} \email jshapiro@ualberta.ca
	\\ \addr Alberta Diabetes Institute, University of Alberta
	}	

\begin{document}
	\maketitle	
	\begin{abstract}
		This paper presents
		advancements in 
		automated early-stage prediction of the success of reprogramming human induced pluripotent stem cells (iPSCs) as a potential source for regenerative cell therapies.
		The minuscule success rate of iPSC-reprogramming of around $ 0.01\% $ to $  0.1\% $ makes it labor-intensive, time-consuming, and exorbitantly expensive to generate a stable iPSC line
		since that requires culturing of millions of cells and intense biological scrutiny of multiple clones to identify a single optimal clone.
		The ability to reliably predict which cells are likely to establish as an optimal iPSC line at an early stage of pluripotency would therefore be ground-breaking in rendering this a practical and cost-effective approach to personalized medicine.
		
		Temporal information about changes in cellular appearance over time is crucial for predicting its future growth
		outcomes.
		In order to generate this data, we first performed continuous time-lapse imaging of iPSCs in culture using an ultra-high resolution microscope.
		We then annotated the locations and identities of cells in late-stage images where reliable manual identification is possible.
		Next, we propagated these labels backwards in time using a semi-automated tracking system to obtain labels for early stages of growth.
		Finally, we used this data to train deep neural networks to perform automatic cell segmentation and classification.
		
		Our code and data are available at ~\url{https://github.com/abhineet123/ipsc_prediction}.
	\end{abstract}


	\section{Introduction}
	\subsection{Motivation}
	The goal of this work is to apply machine learning to automate the identification of
	human iPSCs that show promise for clinical cell therapies in regenerative medicine.
	IPSCs are generated by reprogramming a patient's own cells back in time to make more malleable cells with differentiation potential for generating any cells or tissues of interest.
	This technology has shown great potential for transforming regenerative cell therapies, drug and disease modelling, tissue repair and regeneration, and personalized gene-corrected products.
	However, the pipeline for iPSC generation, characterization and cell banking is a highly labor-intensive, time-consuming and costly one.
	The monetary cost of research-grade iPSC line generation is estimated at USD 10,000-25,000 while that of clinical-grade iPSC line is approximately USD 800,000 based on published reports \citep{Huang2019-vr}.
	The entire process of optimal iPSC line generation and selection can take up to 35 days and requires a further 3 months to produce large scale iPSCs for therapeutic application
	in patients.
	
	Additionally, quality control techniques for growing iPSCs to limit inter- or intra-patient iPSC line variability, which is currently assessed manually, remain imperfect in large-scale biomanufacturing.
	The current solution relies on the judgement of an expert cell biologist, who determines precise iPSC induction, confirms pluripotency based on morphological changes and assesses molecular characterization for multiple clones - all tasks that remain highly effort-intensive and subjectively biased.
	Manual cell quality control therefore cannot be used to scale up the production of iPSCs and derived products for therapeutic applications.
	An automated method enabling high-throughput surveillance and validation of cell identity, growth kinetics, and morphological features is desirable throughout the entire manufacturing process.
	The screening is multifold and needed to not only select optimal cells which have been fully converted to iPSCs during reprogramming stage but also to exclude unstable and pseudo iPSC contaminants during the expansion stage.	
	Automating this process using machine learning would therefore be ground-breaking in improving iPSC bioprocess efficiency and yield, thereby drastically reducing the time and cost involved in the generation of iPSC-based products for therapeutic applications.
	This paper presents some early
	but promising
	steps
	in this direction.
	
	\subsection{Background}
	\subsubsection{iPSC Reprogramming}
	\cite{Takahashi2006-ag} demonstrated that mouse embryonic or adult fibroblasts can be reprogrammed into pluripotent stem cells by introducing four genes encoding transcription factors, namely Oct3/4, Sox2, Klf4, and c-MYC \citep{Takahashi2007-jd,Ye2013-kx}.
	Generated stem cells showed similar morphological or functional behavior as embryonic pluripotent stem cells and were thus termed iPSCs.
	Soon thereafter, \cite{Takahashi2007-jd} reported directed conversion of human fibroblasts into pluripotent stem cells, termed as human iPSCs.
	With the discovery of Yamanaka’s human iPSC technology, patient-derived stem cells have huge potential in regenerative medicine \citep{Takahashi2013-wq}.
	Human iPSCs show merit not only in delivering any desired cell types for treating degenerative diseases, tissue repairing, disease modeling, and drug screening \citep{Amabile2009-pi,Yamanaka2009-fb}, but they also solve two major problems associated with other pluripotent stem cells such as embryonic stem cells \citep{Ye2013-kx}, namely immune tolrenace after transplantation and ethical concerns.	
	However, there still exist technical and biomedical challenges including the risk of teratoma formation and the uncertainty of efficient nuclear reprogramming completeness due to variability and inconsistencies in the selection of optimal cells \citep{Ye2013-kx}.
	There are two major problems to be solved before human iPSCs can be applied as a standardized technique, 
	Firstly, manually monitoring the quality of growing iPSC colonies that is currently practiced does not scale.
	Secondly, only colonies that satisfy clinical good manufacturing practice (GMP) standards need to be identified for use in downstream applications.
	Hence, there is an urgent need for automated quality control, thereby also lending it an element of objectivity and standardization.
		
	\subsubsection{Machine learning in iPSC Recognition}
	Though many applications of machine learning for iPSC recognition in images have been presented in the literature  \citep{Kusumoto2018AutomatedDL,Waisman2019-wv,Zhang2019_xgb,Hirose2021LabelfreeQC,Coronnello2021MovingTI,Kusumoto2022InducedPS,Lan2022MorphologyBasedDL}, there are none that include both detection and classification or use time-lapse imaging, which is the object of this study.
	To the best of our knowledge, \cite{Zhang2019_xgb} presented the method that comes closest to this work though that too differs in several key respects.
	It utilizes fluorescence imaging and the commercial closed-source IMARIS software to segment cells and it captures 3D shape information that is the basis for extracting morphological features to train the  classifier it uses.
	Our aim is to make open-source cell segmentation possible without fluorescence and with only the 2D pixel data in standard phase-contrast microscpy images.
	
	\subsubsection{Deep Learning in Visual Recognition}
	In the past decade, deep learning \citep{Alzubaidi2021_dl_Review} has been applied extensively in computer vision \citep{CHAI2021_cv_review}, especially for recognition tasks like image classification \citep{Byerly2022_classification_review}, object detection \citep{Liu2018_detection_review}, instance segmentation \citep{Gu2022_instance_segmentation_review}, semantic segmentation \citep{Mo2022_semantic_segmentation_review}, and object tracking \citep{Zadeh21_sot_review,Jiao2021_sot_mot_review,Pal2021_mot_review}.
	It has likewise seen broad application in medical image analysis \citep{Suganyadevi2021_Medical_imaging_review,Cai2020_Medical_imaging_review} including segmentation in general \citep{Liu2021_Medical_segmentation_review} and cell segmentation in particular \citep{Wen2022_cell_saegmentation_review}.
	The latter is the task most relevant to this work, though cell tracking \citep{BenHaim2022_cell_tracking_gnn,Chen2021_cell_tracking,Wang2020_cell_tracking_DeepRL,Lugagne20_cell_tracking, Ulman2017_CTC} is also important here.
	More recently, the advent of transformers \citep{Vaswani17_transformer} has led to significant performance improvements \citep{Liu2021_vis_transformer_review} over the convolutional neural network (CNN) based architectures that had been
	prominent
	earlier.
	\cite{Liu2021_Swin_Transformer} proposed the Swin transformer to improve the original vision transformer \citep{Dosovitskiy21_vision_transformer} further using shifted windows.
	This currently appears to be the backbone of choice in most state of the art models, though that might change with the recent introduction of ConvNext \citep{Liu2022_cvnxt} as a competitive CNN-based alternative.
	For this project, we needed instance segmentation models, preferably ones that could benefit from the temporal information in time-lapse images.
	Therefore, we selected top-performing static and video segmentation models (sec. \ref{sec_models}) with publicly available code from a popular leaderboard \citep{PapersWithCode}.
	We also searched through the leaderboards on a couple of cell tracking and segmentation benchmark challenges \citep{Ulman2017_CTC,Anjum2020_CTMC} but failed to find any models with publicly available code.
	
	\section{Methodology}
	
	\subsection{Data Collection}
	\subsubsection{Cell culturing}
	Cells were cultured in a Class-II biocontainment compliant lab with manipulation of cells in a sterile environment using a laminar flow hood with high efficiency particulate air filtration.
	Cells were maintained at 37\degree C with $ 5\% $ $ CO_2 $ within humidified incubators.
	Human iPSCs reprogrammed from patient-derived peripheral blood mononuclear cells (PBMCs) were previously established and characterized for pluripotency in the laboratory and cryostored in ultra-low temperature freezers for long-term storage.
	Before starting the time-lapse imaging, cells from a frozen vial were thawed in a 37\degree C waterbath and cultured on a matrix-coated 6-well plate in iPSC growth media at a seeding density of 100,000 cells per well according to previously published protocol \citep{Park2008-hk}.
	Once the cells got attached, fresh media was replenished in each well every day prior to the time-lapse imaging
	which
	was initiated
	on day 6
	right after cell-seeding.
	
	\subsubsection{Time-lapse imaging}
	Time-lapse images were captured at 15-minute intervals
	on a \href{https://www.microscope.healthcare.nikon.com/en_EU/products/cell-screening/biostudio-t}{Nikon BioStudio-T} microscope using the \href{https://www.microscope.healthcare.nikon.com/products/software/nis-elements}{NIS-Elements} cell observation and image analysis software.
	The images were captured with a 4x lens using a full plate scanning module.
	A total of 275 images were captured spanning 68.75 hours.
	The raw images were $ 10992 \times 10733 $ pixels or 714 megapixels in size though the usable circular cell culture region in the center comprised only about 85 megapixels with a diameter of 5200 pixels.
	A sample image is shown in the supplementary.
	Manual examination showed that images before frame 146 (or 36.5 hours) were unsuitable for our experiments since they contained too many clones, with most being too small and indistinct for reliable labeling.
	Three of the remaining frames - 155, 186 and 189 - were blurry due to camera shake and had to be discarded too.
	As a result, a total of 127 frames were used for all experiments.
	
	\subsubsection{Annotation}
	The original 714 megapixels images are too large to be processed directly so they were first divided into several regions of interest (ROIs) (shown in the supplementary) varying in size from $ 1700 \times 900$ to $ 4333 \times 3833 $.
	These were then annotated in 3 stages.
	\paragraph{Selective Uncategorized Semi-Automated Labeling}
	The 127 frames were first divided into three sets representing different levels of cell development – 146-200, 201-250 and 251-275.
	Set-specific ROI sequences were then created, that is, each ROI spanned only one of the three sets instead of all 127 frames.
	This strategy was chosen to include a good representation of cellular appearance from all stages of development in the labeled data with minimum amount of overall labeling.
	There were 3, 8 and 6 ROIs from the three sets respectively and these 17 ROIs had a total of 656 frames and 4768 cells.
	
	These were then labeled to mark the locations and pixel-wise masks of cells through a custom-designed graphical labeling tool.
	This tool integrates the SiamMask tracker \citep{Wang2018_siammask} to semi-automate the labeling process by propagating manually-created masks
	into future frames by joint unsupervised segmentation and tracking.
	Tracking was stopped manually when it started to fail and then restarted from the last frame where it worked, after making any required fixes in the intermediate frames.
	The labeling tool also supports using a previously trained segmentation model, if available, to automatically generate initial cell candidates that can then be manually modified instead of having to be drawn from scratch, although this capability was not used at this stage.
	Note that this relatively labor-intensive stage did not require involvement from iPSC detection experts since the labeled cells were not categorized into good and bad.
	
	\paragraph{Exhaustive Automated Labeling}
	A Swin transformer instance segmentation model \citep{Liu2021_Swin_Transformer} was first trained on the annotations from the previous stage.
	Next, ROI sequences spanning all 127 frames were created.
	These were designed to cover as much of the circular well area containing cells as possible while minimizing overlap between different ROIs. 
	A total of 31 sequences were created by extending 11 of the 17 ROIs from the previous stage to the remaining frames and creating 20 new ones.
	These sequences had 3937 frames with 22,598 cells in all.
	More details about this dataset as well as the one from the previous stage are available in the supplementary.
	Finally, the trained Swin transformer model was used to automatically detect and segment cells in each of these frames.
	
	\paragraph{Categorized Retrospective Labeling}
	\label{retrospective}
	An iPSC detection expert first manually categorized the cells in frame 275 from each of the 31 ROIs into good and bad.
	The two categories were respectively named iPSC and differentiating cell (\textbf{DfC}) to reflect their likely future growth outcomes.
	A semi-automated interactive multi-object tracking tool was then used to propagate these labels backwards in time by tracking each cell line from frame 275 to 146.
	This process accounted for cell division and fusion events
	\footnote{Note that a division event when going backwards in time is actually a fusion event and vice versa}
	by giving each child cell the same label as the parent in case of division and requiring that all merging cells have the same label in case of fusion.
	A violation of the latter requirement would have meant that the human expert's labels were incorrect but this never happened
	which is a sign of their reliability.
	Note that cell lines that disappear before reaching frame 275 cannot be
	categorized
	in this way
	so these have been excluded from all experiments.
	
	The tracking algorithm was kept simple due to time and computational constraints.
	Cells were associated between neighbouring frames on the basis of location and shape, similar to the IOU tracker \citep{Bochinski17_iou} while
	likely fusion and division events were detected using heuristics based on the extent of change in the size, shape and location of associated cells, in addition to association failures.
	The detailed algorithm is included in the supplementary.
	Due to this simplicity, the retrospective labelling process is currently more time- and labor-intensive than ideal and it took between 10 and 30 minutes to label each ROI sequence depending on density of cells and frequency of division and fusion events.
	
	We have made the code for all three stages publicly available \citep{ipsc_pred_github} along with the annotated data and trained models to facilitate reproducibility of our results and further work in this domain.
	
	\subsection{Models}
	\label{sec_models}
	We selected two state-of-the-art image classification models to allow head-to-head comparison with the XGBoost classifier-based approach by \cite{Zhang2019_xgb} (\textbf{XGB}) which is the only existing method in literature that is relevant to this work.
	The models we chose are based on Swin transformer \citep{Liu2021_Swin_Transformer} (\textbf{SWC}) and ConvNext \citep{Liu2022_cvnxt} (\textbf{CNC}) architectures.
	We also selected five instance segmentation models since both cell detection and classification are needed in the absence of fluorescence that was used by \cite{Zhang2019_xgb} to identify cells.
	Two of these are static detectors that process each frame independently and discard any video information. 
	They are both variants of Cascade Mask RCNN \citep{Cai19_cascade_rcnn} that differ in their backbone architectures, these being the same as the classifiers chosen above - Swin transformer \citep{Liu2021_Swin_Transformer} (\textbf{SWD}) and ConvNext \citep{Liu2022_cvnxt} (\textbf{CND}).
	The remaining three are video detectors that combine information from multiple video frames to make their decisions.
	One of these - \textbf{IDOL} \citep{Wu2022_idol} - is an online model that only uses information from past frames while the other two - SeqFormer \citep{Wu2021_SeqFormer} (\textbf{SEQ}) and \textbf{VITA} \citep{heo2022_vita} - are batch models that use information from the entire video sequence including both past and future frames.
	All three video detectors
	use versions of the Swin transformer backbone.
	We also experimented with two ResNet \citep{He16_resnet} variants of VITA but they did not perform well and so have been excluded here.
	In addition, we trained a Swin transformer semantic segmentation model and tried several ensemble techniques to combine semantic and instance segmentation results but these did not yield significant performance improvements and are therefore likewise excluded.
	
	\subsection{Training}
	\label{training}
	Since retrospective labeling is the most time-consuming part of our pipeline, it is desirable to minimize the need to label backwards to as few frames as possible.
	Therefore, in addition to comparing modern deep learning-based methods with XGB, we also wanted to evaluate the extent to which  model performance on early-stage images depends on the lateness of the frames on which it is trained.
	We constructed two different training datasets to achieve this - an early-stage set with 38 frames from 163 to 202 and a late-stage set with 73 frames from 203 to 275.	
	Models trained on both datasets were evaluated on the 16 frames from 146 to 162.
	We also had to adapt the method proposed by \cite{Zhang2019_xgb} to work with our images since that work computes several of the features using 3D image information which is not available in our case.
	It turned out that only 7 of the 11 features used there could be suitably approximated with 2D data so we trained XGB using only these 7 features whose details are in the supplementary.
	All models were trained to convergence using mostly default settings recommended for each model by the respective authors except for minor tinkering with batch sizes, data augmentation strategies and scaling factors to make the models fit in the limited GPU memory available.
	The classification datasets were constructed from image patches corresponding to the bounding box around each labeled cell with an additional 5 pixel border for context.
	Models were trained on several different GPU configurations -
	video detectors: 2 $ \times $ Tesla A100 40 GB,
	static detectors: 2 $ \times $ RTX 3090 24 GB,
	classifiers: 3 $ \times $ RTX 3060 12 GB and 3 $ \times $ GTX 1080 Ti 11 GB.
	
	\section{Evaluation}	
	\subsection{Classification Metrics}
	\label{cls_metrics}
	We used standard receiver-operating-characteristic (ROC) curves and the corresponding area-under-curve (AUC) metric to compare the models.
	Note that head-to-head comparison between detectors and classifiers is difficult since the former both detect and classify cells while the latter only do classification on all the cells in the ground truth (GT).
	Two types of detector failures need to be accounted for in order to render such a comparison meaningful:
	\begin{itemize}[noitemsep,topsep=0pt]
		\item False Positives (FP): Detections without matching GT cells
		\begin{itemize}[left=0pt,topsep=0pt,noitemsep,label=\textendash]
			\item Misclassification (\textbf{FP-CLS}): a GT DfC is detected but misclassified as iPSC
			\item Duplicates (\textbf{FP-DUP}): the same GT iPSC is detected multiple times and classified each time as iPSC		
			\item Non-Existent (\textbf{FP-NEX}): an iPSC is detected where no GT cell exists (neither iPSC nor DfC)		
		\end{itemize}
		\item False Negatives (FN): GT cells without matching detections
		\begin{itemize}[left=0pt,topsep=0pt,noitemsep,label=\textendash]
			\item Misclassification (\textbf{FN-CLS}): a GT iPSC is detected but misclassified as DfC		
			\item Missing detection (\textbf{FN-DET}): a GT iPSC is not detected at all (neither as iPSC nor as DfC)
		\end{itemize}
	\end{itemize}
	FP-NEXs were ignored when computing the classification metrics since manual examination\footnote{visualizations for
		all
		detection failures are included in the supplementary material}
	showed that
	virtually
	all of these corresponded to
	one of
	two cases, neither of which is important in our application:
	\begin{itemize}[noitemsep,topsep=0pt]
		\item \textbf{FP-NEX-WHOLE}: unlabeled cells whose labels could not be inferred by retrospective labeling (sec. \ref{retrospective})
		\item \textbf{FP-NEX-PART}: parts of labeled cells mostly
		involving ambiguously-shaped cells that could plausibly be interpreted as undergoing division or fusion
		but were labeled as whole cells.
	\end{itemize}
	FP-DUPs were also ignored since multiple detections of an iPSC have little impact in our application.
	Finally, FN-DETs had to be discarded since ROC curves can only be generated by varying the cut-off used in
	filtering detections based on their confidence values
	which
	are unavailable for
	undetected
	cells.
	
	In addition to the AUC of the complete ROC curve, we also used partial AUCs \citep{McClish1989_partial_auc} with FP thresholds of $ 0.1\% $, $ 1\% $ and $ 10\% $.
	Even a small number of FPs can be extremely detrimental in our application due to the high cost of culturing non-viable cells so that a model that performs better at these
	FP rates is preferable to another that is better overall but underperforms here.

	\subsection{Detection Metrics}
	\label{det_metrics}
	The exclusion of FN-DETs while computing the classification metrics can make these
	biased in favour of detectors with high rates of missing cells but high
	accuracy for the few cells that they do detect.
	We accounted for this by incorporating the following detection metrics to evaluate and compare only the detectors:
	\begin{itemize}[noitemsep,topsep=0pt]
		\item Frequency of FN-DETs, FP-DUPs and FP-NEXs
		\item Standard detection metrics \citep{Huang2016_SpeedAccuracyTF} of average precision (\textbf{AP}) and AUC of the recall-precision curve (\textbf{RP-AUC}) 
	\end{itemize}
	\subsection{Temporal Metrics}
	\label{temporal_metrics}
	The ability to detect iPSCs as early as possible
	is crucial to our application.
	Therefore, we also evaluated the models in each of the 16 test frames one-by-one to judge how their performance varied over time.  
	A model that performs better in earlier frames would be preferable to one that performs better overall but underperforms in earlier frames.
	\subsubsection{Subsequenctial Inference}
	Video detectors detect cells spanning all the frames in their input video instead of frame-by-frame which is incompatible with temporal evaluation since even detections in early frames are done using information from all 16 frames in the test set.
	To resolve this, we used incremental inference where the detections for each frame are generated by running the detector on a subsequence comprising only that frame and all of the preceding ones so that information from future frames is not used.
	For example, detections for frames 1, 2 and 3 are respectively generated by running the detector only on subsequences comprising frames (1), (1, 2) and (1, 2, 3).
	
	We used another variant of
	subsequential inference to evaluate the impact of the number of frames on the performance of video detectors so we can judge whether
	patterns in the way that cell boundaries change over time provide useful information about their eventual outcome to these detectors.
	Here, we divided the 16-frame sequence into a set of non-overlapping subsequences, each with a fixed size, and ran inference on each subsequence independently.
	For example, with a subsequence size of 2, our 16-frame sequence is divided into 8 subsequences
	- (1, 2), (3, 4), ..., (15, 16).
	Detections for all frames in each subsequence are then generated by running on only the frames in that subsequence.
	We experimented with subsequence sizes of 1, 2, 4 and 8.
	
	\begin{figure*}[t]
		\includegraphics[width=0.495\textwidth]{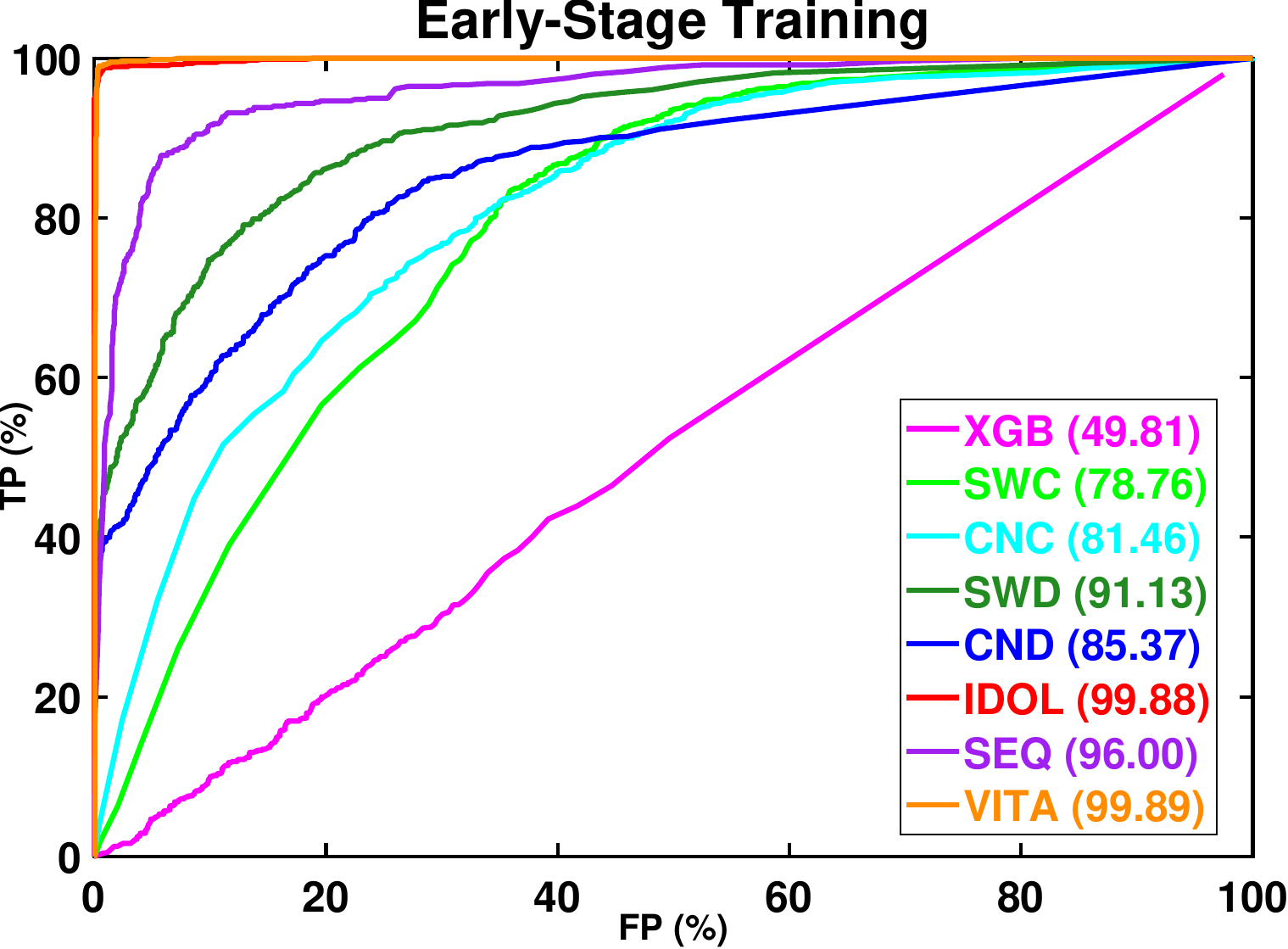}
		\includegraphics[width=0.495\textwidth]{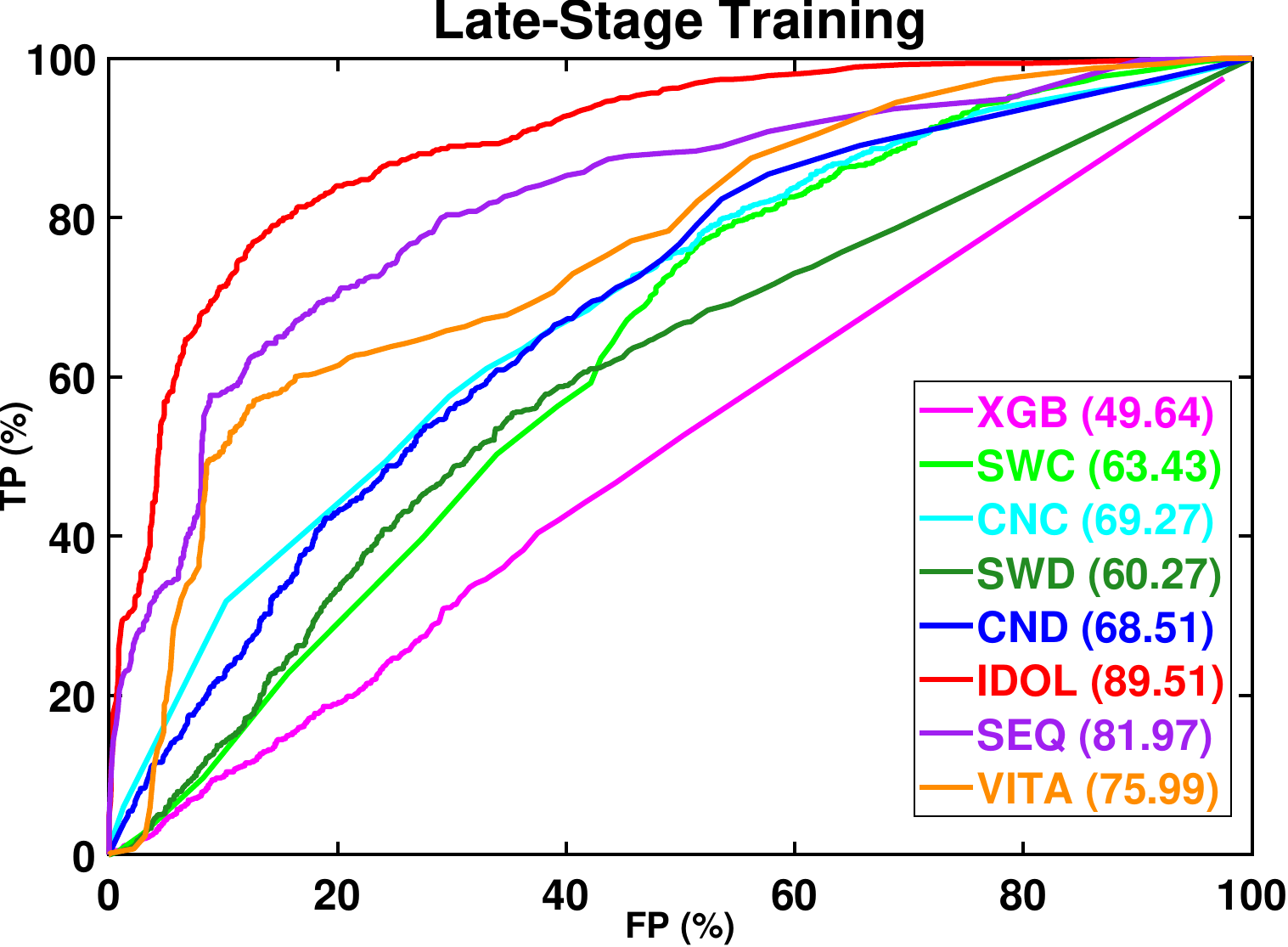}		
		\includegraphics[width=0.495\textwidth]{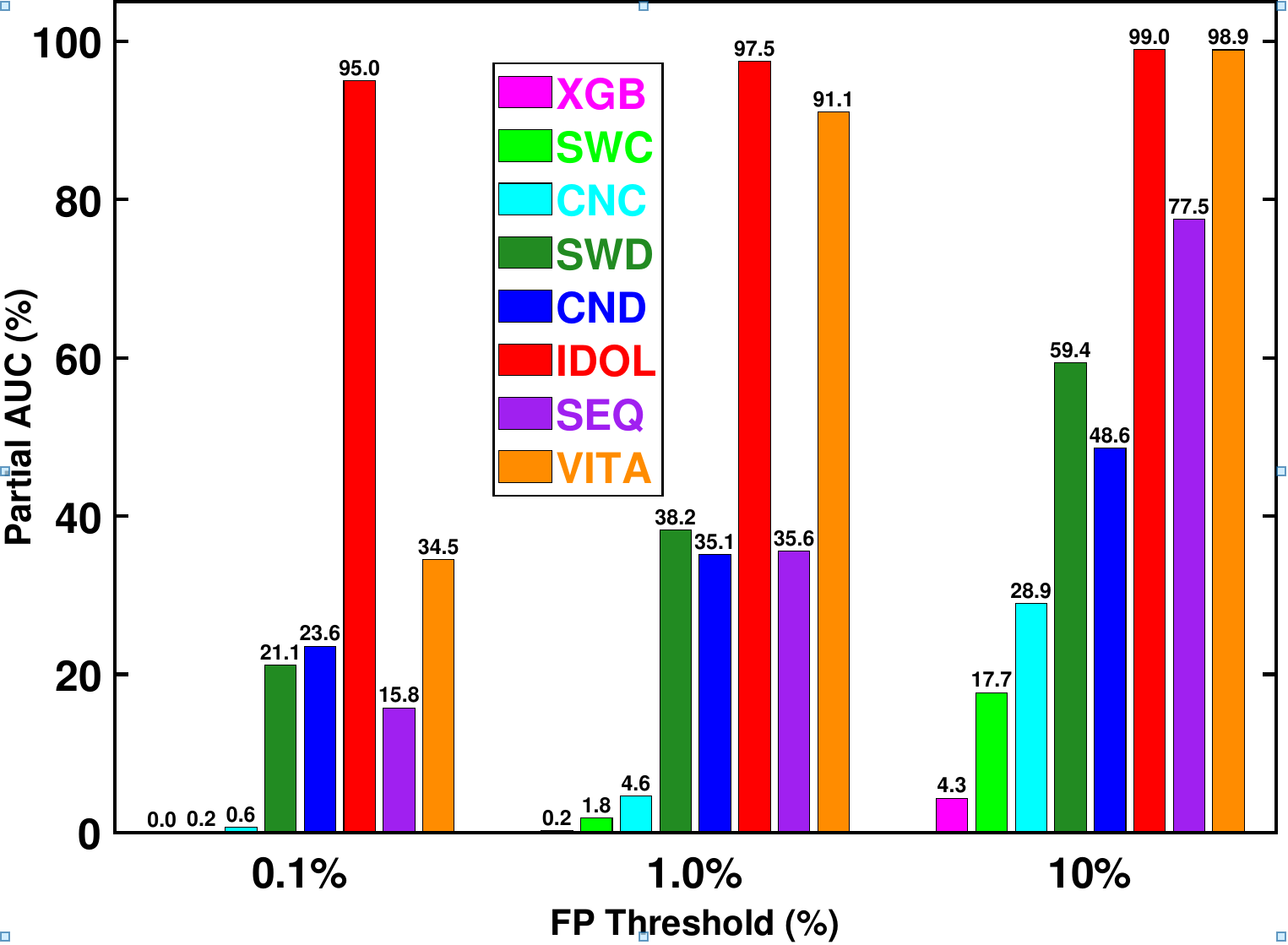}		
		\includegraphics[width=0.495\textwidth]{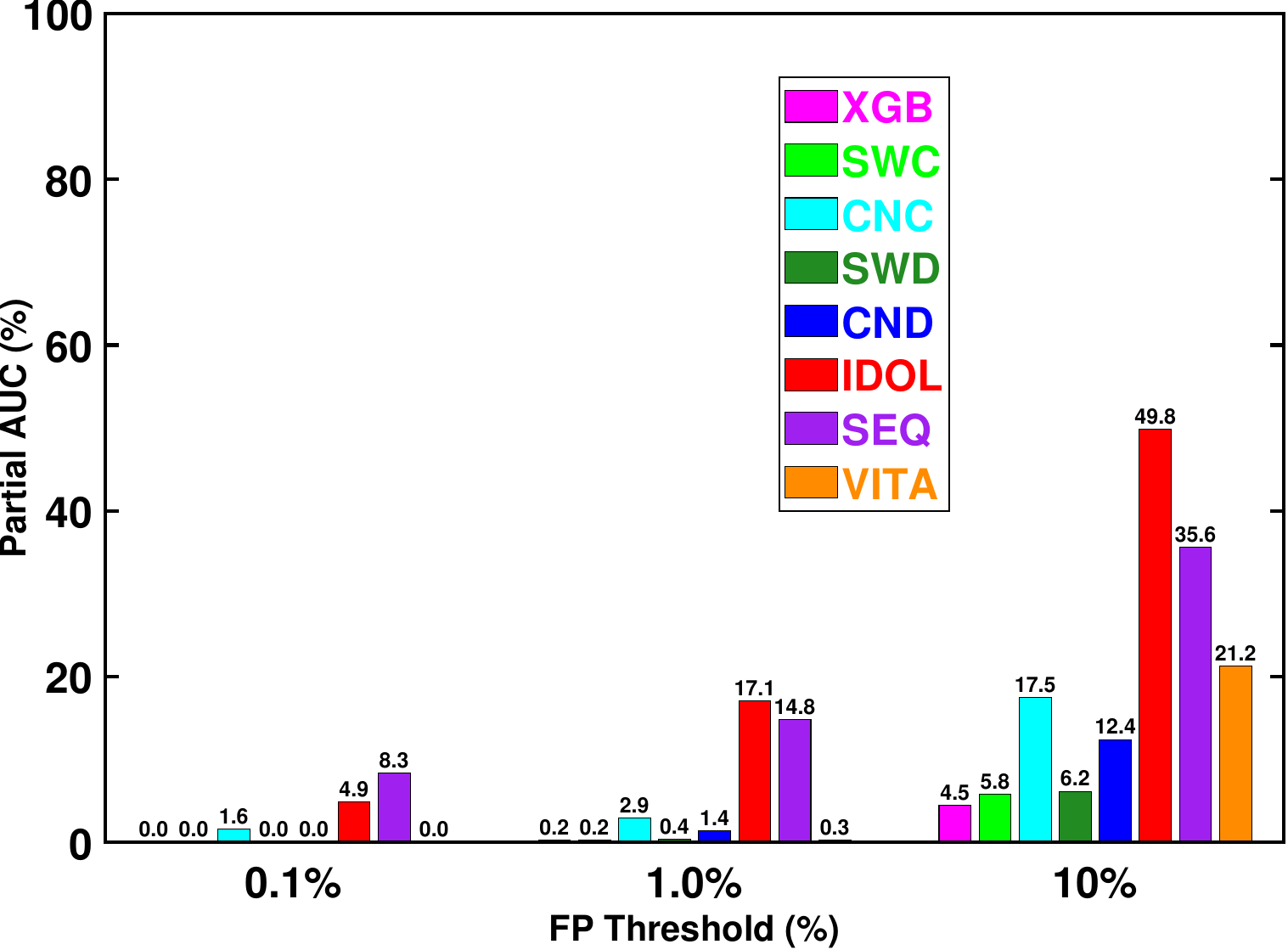}
		\caption{
			Classification metrics for
			(left) early and (right) late-stage models.		
			Top: ROC curves (respective AUC values are in the legend);
			bottom: partial AUCs.
			Please refer sec. \ref{sec_models} for model acronyms.
		}
		\label{fig_cls}	
	\end{figure*}
		
	\subsection{Results}	
	\subsubsection{Classification}	
	ROC curves for both early and late-stage models are shown in the top row of Fig. \ref{fig_cls}.
	It can be seen that the early models perform much better than the late ones, which is expected since their training images are much more similar to the test images.
	However, the extent of this difference does indicate that cell-appearance changes so much over the course of just 9.5 hours (temporal gap between early and late-stage training images) that
	long-term retrospective labelling right to the very early-stage images
	is likely to be essential
	to generate training data for models that can do early-stage iPSC detection reliably.
	The significant performance advantage of deep learning models over XGB is also apparent here.
	The main shortcoming of XGB seemed to be its inability to handle class imbalance in the training set.
	Since about 75\% of all cells were DfCs, XGB apparently learnt to classifty nearly all cells as DfCs.
	We tried all standard techniques to handle class imbalance \citep{Krawczyk2016_imbalance} but this is the best performance we could get from it.
	
	Further, all the video detectors are consistently better than the static ones which
	seems
	to confirm the human experts' supposition that temporal information is crucial for making good predictions.
	Also, the static detectors do outperform the classifiers but only significantly so in the early-stage case.
	Since the precise shape of cell boundaries is not available to the classifiers, this
	lends some weight
	to the additional supposition that cell-shape is important for recognizing iPSCs.
	However, as already noted above, the cell-shapes change too rapidly for this information to be generalizable from the later stages to the earlier ones.
	Finally, IDOL turns out to be the best model overall even though it is the smallest and fastest of the three video detectors while the much larger VITA shows a susceptibility to overfitting
	in its sharp decline between the two cases.	
	This kind of overfitting is exhibited by both the static detectors too, though it is more strongly marked in case of Swin transformer.
	
	Partial AUCs are shown in the bottom row of Fig. \ref{fig_cls}.
	Relative performance between the models is broadly similar to that for overall AUC though IDOL shows greater performance advantage over other models in the early-stage case especially for $  0.1\% $ FP.
	The detectors also show greater improvement over the classifiers in these high-precision scenarios.
	We also analysed the temporal evolution of partial AUCs by evaluating these frame-by-frame to generate 3D plots with time dimension on the Z-axis (included in the supplementary material) but could not find any useful patterns beyond those apparent in these 2D plots.
	\subsubsection{Detection}	
	\begin{figure*}[t]
		\includegraphics[width=0.495\textwidth]{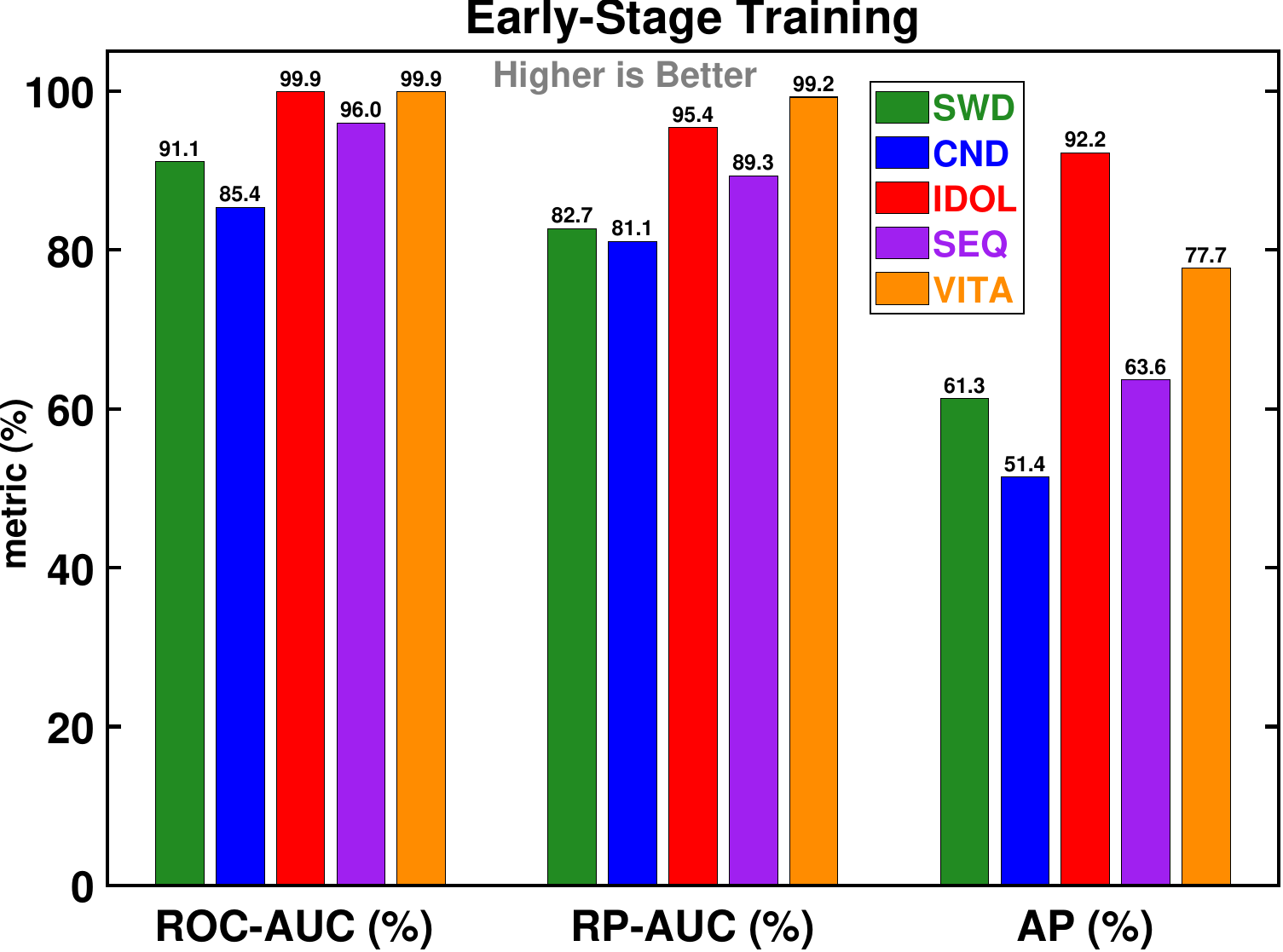}
		\includegraphics[width=0.495\textwidth]{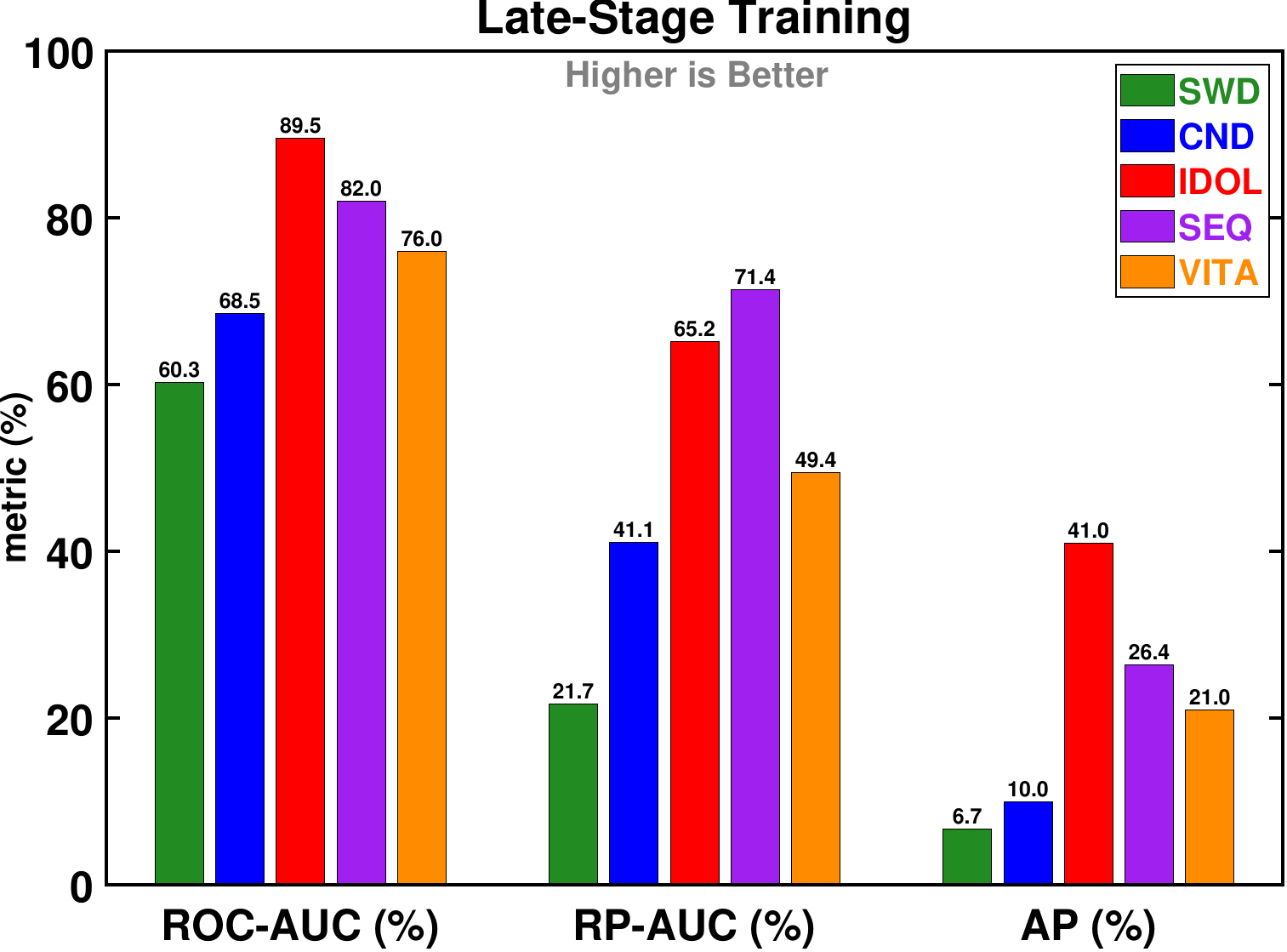}		
		\includegraphics[width=0.495\textwidth]{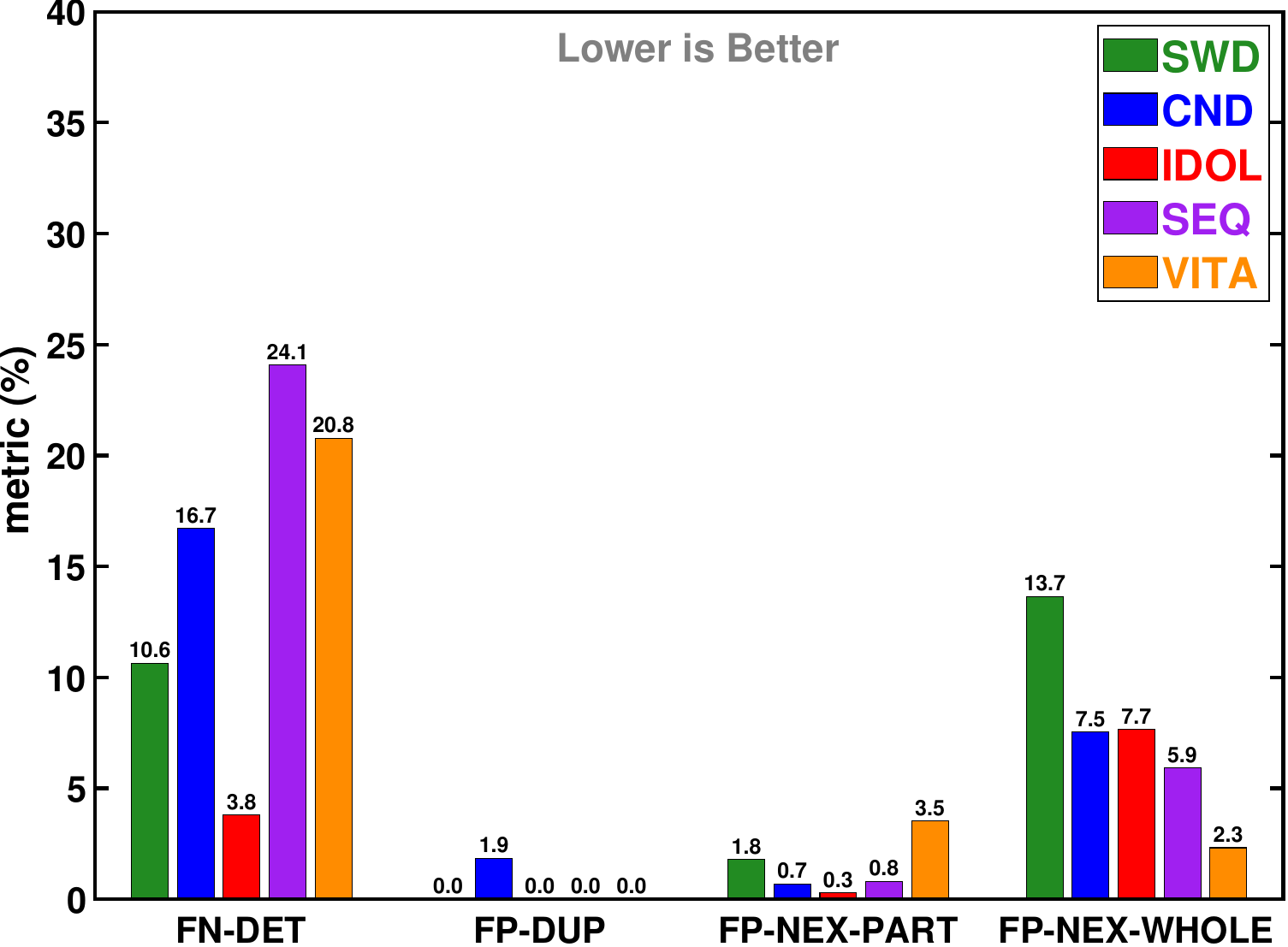}		
		\includegraphics[width=0.495\textwidth]{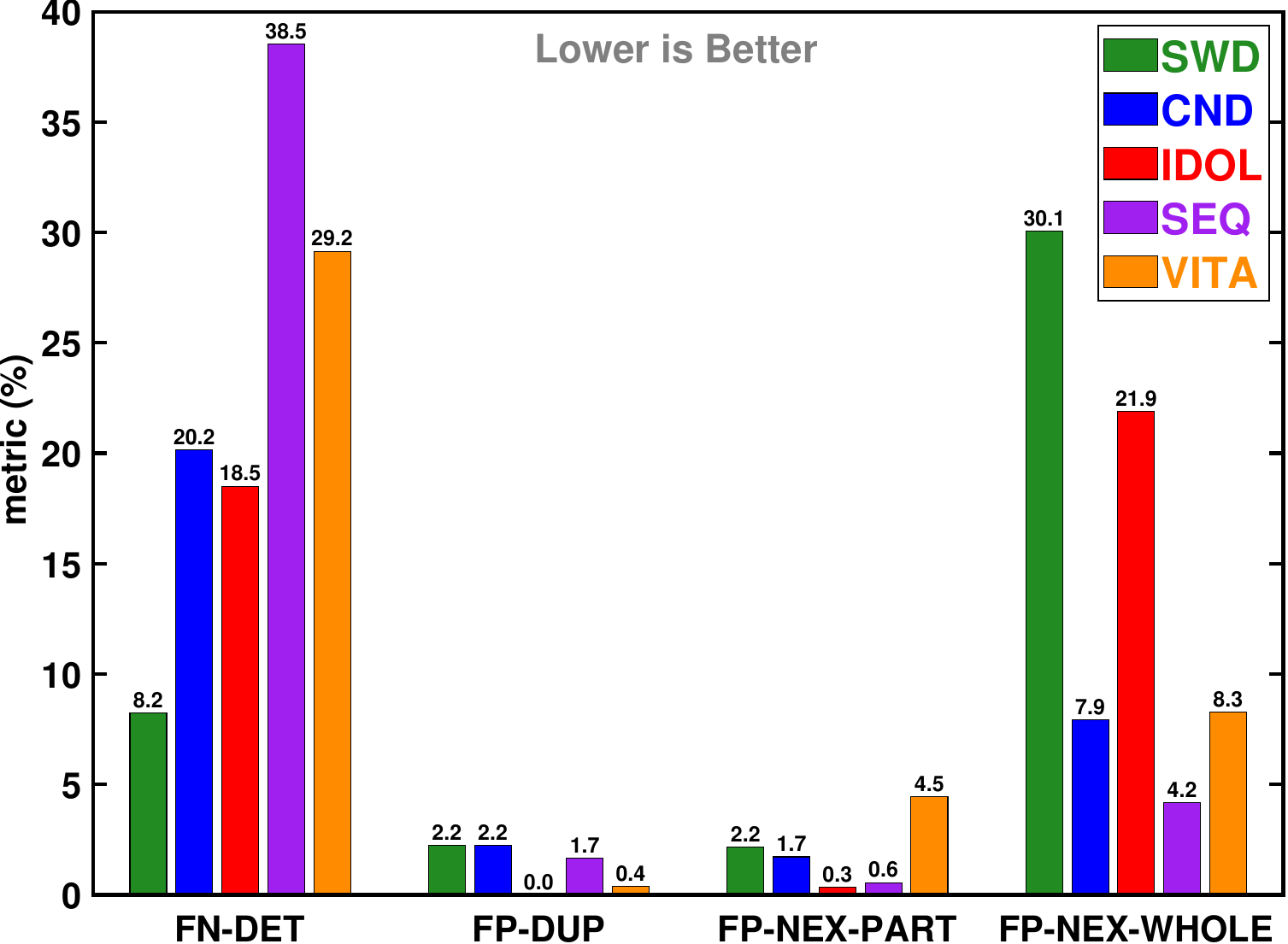}
		\caption{
			Detection metrics
			for (left) early and (right) late-stage models.
			ROC-AUC is included for comparison.
		}
		\label{fig_det}	
	\end{figure*}
	As shown in Fig. \ref{fig_det}, relative detection performance is mostly consistent with classification accuracy, with IDOL still being the best model overall.
	IDOL also shows the smallest drop in performance between ROC-AUC and AP while the two static detectors show the largest drops amounting to nearly 10-fold and 7-fold respectively for SWD and CND in the late-stage case. 
	Among the video detectors, VITA has the largest drop, especially in the late-stage case which underlines its overfitting.
	Further, both SEQ and
	VITA have very high FN-DET rates that are also at odds with their relatively good classification performance.
	Conversely, SWD has much lower rates of FN-DETs than its ROC-AUCs would suggest, especially in the late-stage case where it outperforms all other models
	by a significant margin, though this does come at the cost of a corresponding rise in FP rates.
	We can also note that this increase in FPs
	is dominated by only one subtype, namely FP-NEX-WHOLE, while FP-DUP and FP-NEX-PART show little change, not only for SWD but also for IDOL and VITA.
	It appears that either due to the greater range of cell-appearances in the late-stage dataset (owing to its greater size) or larger disparity in cell-appearance with respect to the test set, these models learnt to detect a lot more of the unlabeled cells than their early-stage counterparts, which were therefore better at discriminating between the unlabeled and labeled cells.
	\subsubsection{Temporal}	
	\begin{figure*}[t]
		\includegraphics[width=0.495\textwidth]{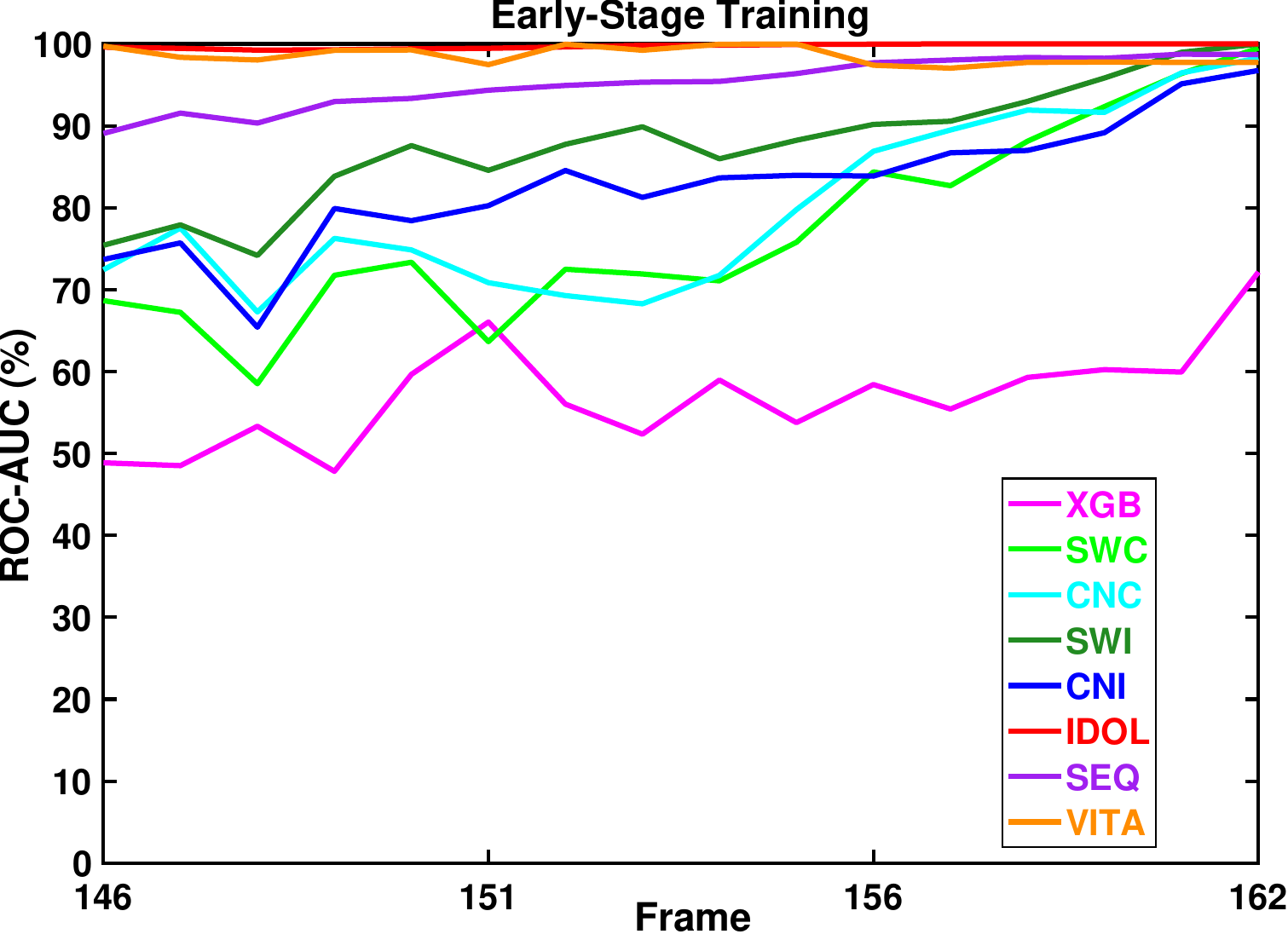}
		\includegraphics[width=0.495\textwidth]{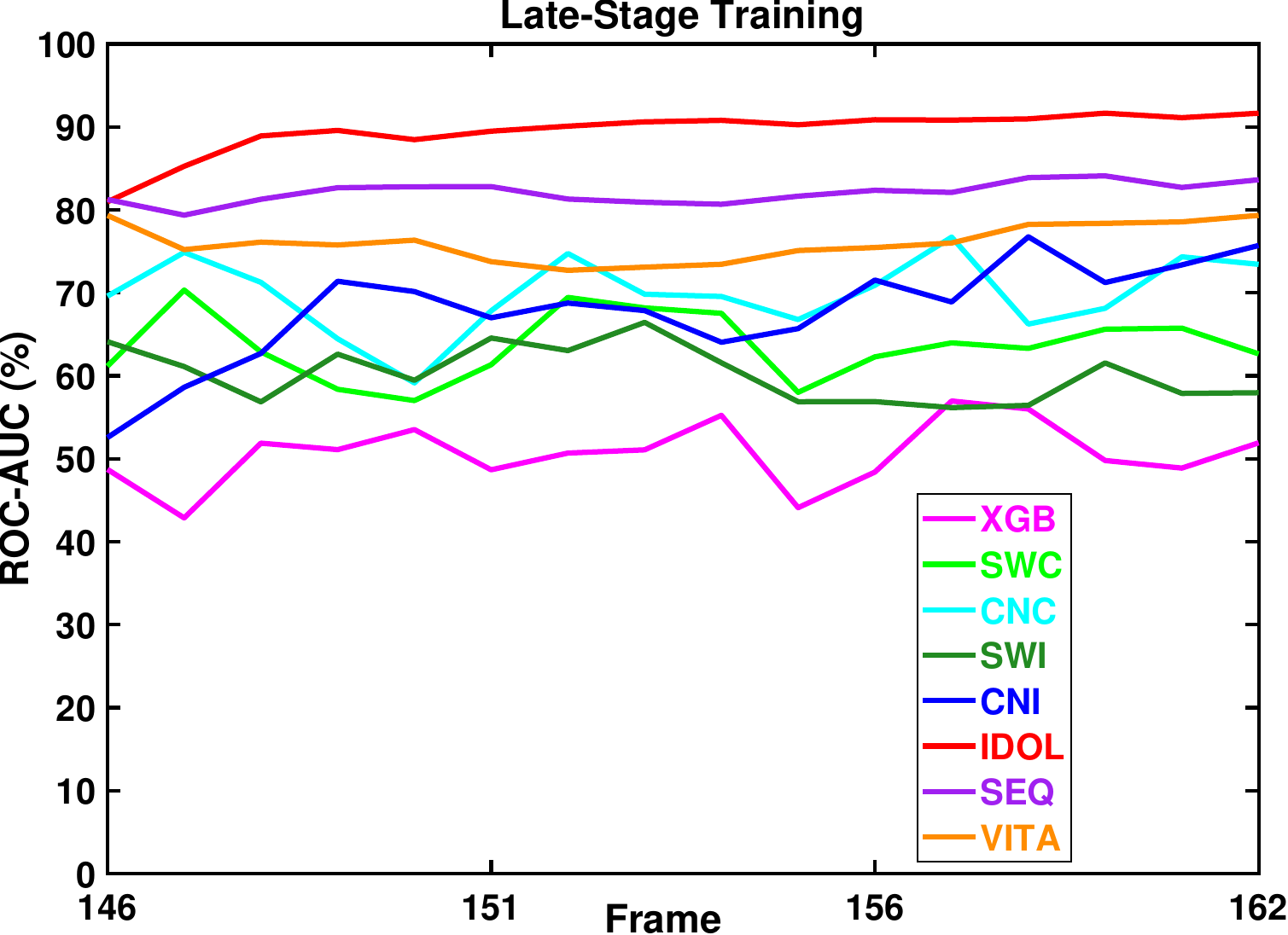}	
		\includegraphics[width=0.328\textwidth]{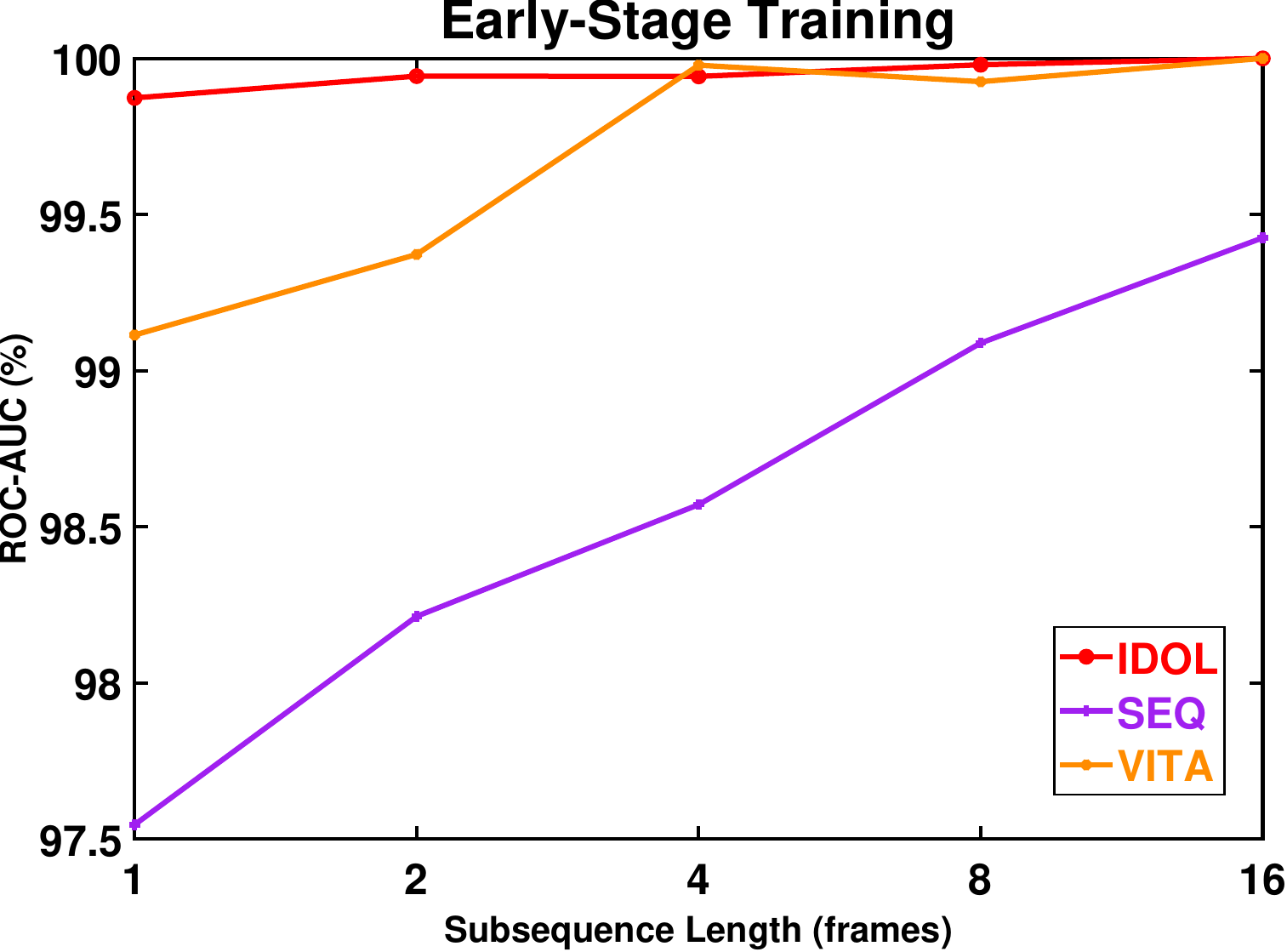}
		\includegraphics[width=0.328\textwidth]{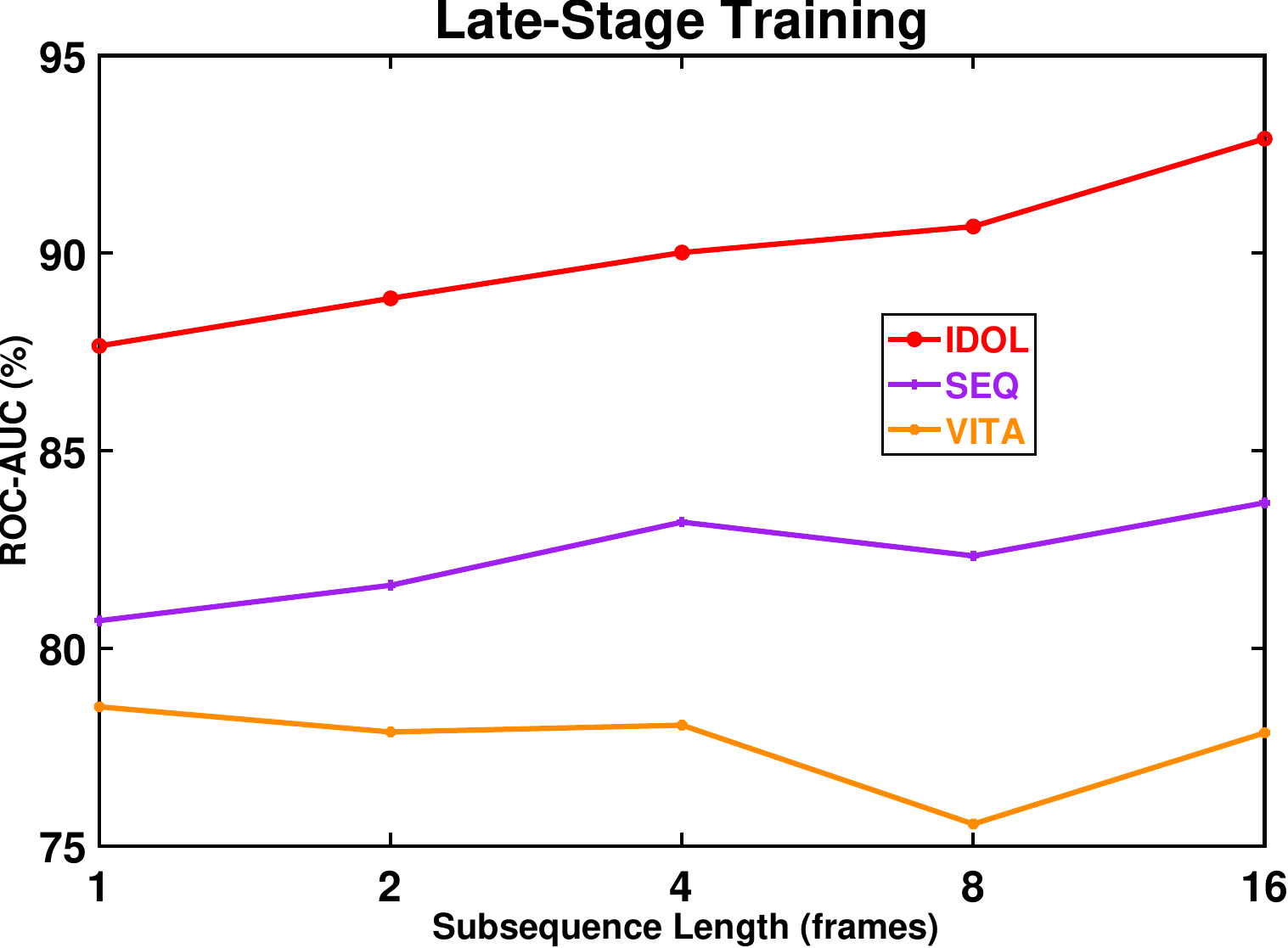}
		\includegraphics[width=0.328\textwidth]{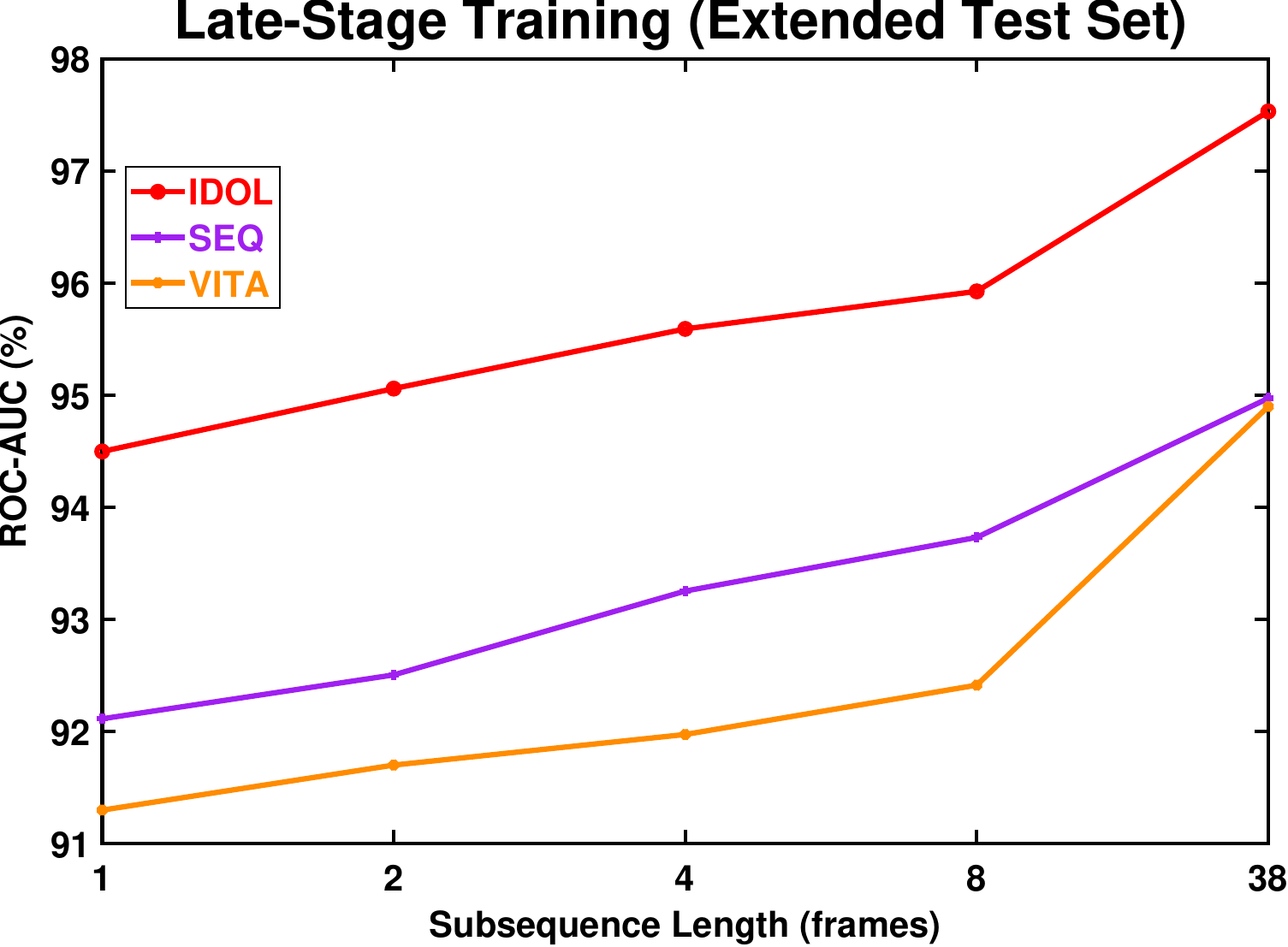}
		\caption{
			Temporal metrics - 
			top: frame-wise ROC-AUCs for (left) early and (right) late-stage models;
			bottom: subsequential ROC-AUCs for video detectors - left and center plots show early and late-stage models tested on the standard test set of frames 146 - 162 while the right one shows the latter tested on an extended test set with frames 146 - 201.
			Note the curtailed and variable Y-axis ranges in the subsequential ROC-AUC plots.
		}
		\label{fig_temporal}	
	\end{figure*}
	Frame-wise AUCs are shown in the top row of Fig. \ref{fig_temporal}.
	Somewhat contrary to expectation, AUC does not show consistent increase with time even though test images are
	becoming more similar to the training images.
	In fact, many of the early-stage models do show a weak upward trend while late-stage ones do not show any.
	This is particularly unexpected since the latter perform significantly worse overall and show signs of overfitting, which should lead to a more strongly marked increase in accuracy as resemblance between the test and training images increases.
	A possible explanation might be that the late-stage training images are just too far away from the test set for any intra-set variations in test images to make enough of a difference in their resemblance to training images to benefit the models.
	
	Bottom row of Fig. \ref{fig_temporal} shows the impact of subsequence length on the
	video detectors.
	These plots include only ROC-AUC but detection metrics showed similar patterns and have thus been relegated to the supplementary.
	In order to incorporate the longest-term temporal information possible with our dataset, we also tested the late-stage model on an extended test set comprising frames 146 to 201.
	Unfortunately, there is a visual discontinuity between frames 184 and 185 due to which this 54-frame sequence had to be divided into two subsequences -  one with 38 frames from 146 to 184 and another with 16 frames from 185 to 201 - so that the longest subsequence length was 38 instead of 54.
	There is indeed a general upward trend with subsequence length but it is much weaker than might be expected.
	Early-stage models showed negligible impact of subsequence length while the greatest overall gains were $ 5.2\% $ in the second case and $3.6\% $ in the third case, achieved respectively by IDOL and VITA.
	Neither of these seem sufficient considering the 4 to 9 hours worth of extra temporal information available to the models.
	This might indicate that changes in cell-appearance over time are not as useful for recognizing iPSCs as supposed by experts.
	More likely, it might mean that existing video detectors are simply not good enough to exploit this information sufficiently well and better long-term models are needed.
	\section{Conclusions}
	This paper presented a labeling, training and evaluation pipeline
	along with baseline performance results for early-stage prediction of iPSC reprogramming outcome to help select the best quality clones.
	These are still early days of research in this domain and it is difficult to say how practical such an automation can be.
	While it is clear that deep learning models hold significant advantage over previous methods, they also show signs of overfitting and it is unclear how much training data would be needed to overcome such issues.
	
	\section{Future Work}	
	We are currently working to improve the tracking algorithm in the retrospective labeling system to make the process faster and less tedious since that is the current bottleneck in the labelling pipeline.
	We are trying out several state-of-the-art real-time multi object trackers, especially those specialized for tracking cells \citep{Chen2021_cell_tracking,Wang2020_cell_tracking_DeepRL,Lugagne20_cell_tracking}, to replace the simple IOU based algorithm used in this work.	
	We are also looking into ways to improve this algorithm by exploiting recursive parent-child relationships in our labels that can be used to construct hierarchical tree-like structures.
	Transformer architectures that support such structures \citep{Wang2019_TreeTI,Ahmed2021_EncodingDI,Wang2022_ATT,Zhong2022_ATS} might help to incorporate cell division and fusion events directly into the algorithm to not only improve the cell association reliability but also reduce the incidence of false positives in the detection of these events, which is the most time-consuming and tedious aspect of this process.
	A recent method uses graph neural networks to exploit these structures \citep{BenHaim2022_cell_tracking_gnn} for cell tracking.
	We are trying to incorporate it into our pipeline and also improve it further using transformers.	
	If these efforts prove successful, we hope to make long term retrospective labelling spanning multiple weeks feasible so any such demands for data can be met.
	
	We are also exploring ways to better exploit
	long-term patterns in the evolution of cell-appearance over time
	to improve prediction quality
	since the current ability of video detectors in this regard does not appear to commensurate 
	with the importance that human experts attach to this information.
	This discrepancy might be due to both the limited temporal span of our data or the models being unable to benefit from this information because they do not learn from sufficiently wide temporal windows.
	We hope to resolve the former by extending our data from its current span of only 3 days to as much as 5 weeks once the retrospective labeling system have been improved enough to make such long-term labeling feasible.
	The latter can be addressed by using more memory-efficient models since increasing the size of temporal windows for training is an extremely memory expensive process and our computational resources are insufficient to train existing models with wider windows.
	
	Another related issue we are trying to address is that of the black-box nature of deep learning which makes it very difficult for the medical experts in our team to figure out the reasons behind particular failures of the models.
	It makes it equally difficult for us to incorporate their suggestions to fix such failures.
	We are hoping that recent advances in the field of interpretable deep learning \citep{Li2021_interpretable_dl_review,Samek2021_interpretable_dl_review,molnar2020_interpretable_dl_book,samek2020_interpretable_dl_review,samek2019_interpretable_dl_book}, especially with regard to transformers \citep{Ali22_interpretable_dl_transformer}, video processing \citep{anders2019_interpretable_dl_video} and medical imaging \citep{brima2022_interpretable_dl_medical}, might help with these limitations.

	
	\acks{Fig. 1 in the graphical abstract is provided by Zofia Czarnecka. Figure created using Biorender.com}
	
	%
	\ethics{The work follows appropriate ethical standards in conducting research and writing the manuscript, following all applicable laws and regulations regarding treatment of animals or human subjects.}
	
	\coi{We declare we don't have conflicts of interest.}

	\bibliography{ipsc_references}
	
\end{document}